\documentclass{article}

\usepackage{ijcai13}

\newtheorem {definition}{Definition}

\newtheorem {theorem}{Theorem}

\newtheorem {example}{Example}

\usepackage{amsmath}
\usepackage{amssymb}

\usepackage{latexsym}
\usepackage{picinpar}

\begin{document}

\title{Logical Fuzzy Optimization}

\author{ Emad Saad \\
%Institute for Artificial Intelligence and Biological Systems \\
%School of Computing \\
%University of Leeds \\
%LEEDS \\
%United Kingdom \\
emsaad@gmail.com
}

\maketitle

\begin{abstract} We present a logical framework to represent and reason about fuzzy optimization problems based on fuzzy answer set optimization programming. This is accomplished by allowing fuzzy optimization aggregates, e.g., minimum and maximum in the language of fuzzy answer set optimization programming to allow minimization or maximization of some desired criteria under fuzzy environments. We show the application of the proposed logical fuzzy optimization framework under the fuzzy answer set optimization programming to the fuzzy water allocation optimization problem.

\end{abstract}

\section{Introduction}

Fuzzy answer set optimization is a logical framework aims to solve optimization problems in fuzzy environments. It has been shown that many interesting problems including representing and reasoning about quantitative and qualitative preferences in fuzzy environments and fuzzy optimization can be represented and solved using fuzzy answer set optimization. This has been illustrated by applying fuzzy answer set optimization to the course scheduling with fuzzy preferences problem \cite{Saad_FuzzPref}, where instructor preferences over courses are represented as a fuzzy set over courses, instructor preferences over class rooms are represented as a fuzzy set over class rooms, and instructor preferences over time slots are represented as a fuzzy set over time slots. The course scheduling with fuzzy preferences problem \cite{Saad_FuzzPref} is a {\em fuzzy optimization} problem that aims to find the optimum course assignments that meets all the instructors top fuzzy preferences in courses, class rooms, and time slots. Moreover, it has been shown in \cite{Saad_FuzzPref} that fuzzy answer set optimization can be used to solve both {\em crisp} optimization problems and fuzzy optimization problems in a unified logical framework .

However, the lack of fuzzy aggregates preferences, e.g., minimum and maximum, in fuzzy answer set optimization makes the framework less suitable for representing and solving some fuzzy optimization problems that are based on minimization and maximization of some desired criteria imposed by the problem. For example, consider the following fuzzy optimization problem from \cite{UNESCO}.

\begin{example} Assume that we want to find the water allocation for each of the three firms, which are located along a river, in a way that maximizes the total benefits of the three firms. Consider $x_1$, $x_2$, $x_3$ are the units of water allocation to firms one, two, and three respectively. Consider also that the benefits of the three firms denoted by $B_1$, $B_2$, and $B_3$ respectively are given by $B_1 = 6x_1 - x^2_1$, $B_2 = 7x_2 - 1.5x^2_2$, and $B_3 = 8x_3 - 0.5x^2_3$. The water allocations cannot exceed the amount of water available in the river minus the amount of water that must remain in the river. Assume that amount is $6$ units. The target is to maximize the total benefits, $T(X)$, the objective function, which is
\[
maximize \; T(X) = (6x_1 - x^2_1) + (7x_2 - 1.5x^2_2) + (8x_3 - 0.5x^2_3)
\]
subject to $x_1 + x_2 + x_3 \leq 6$.
\\
\\
However, the set of possible values of $T(X)$ are not precisely defined, rather each possible value of $T(X)$, for $X = (x_1, x_2, x_3)$, is known to some degree, where the higher the value of $T(X)$ the higher the degree of $T(X)$. The degree of each value of $T(X)$ is given by the fuzzy membership function ({\em objective membership function}),
\[
D_g(X) = \frac{(6x_1 - x^2_1) + (7x_2 - 1.5x^2_2) + (8x_3 - 0.5x^2_3)} {49.17}
\]
%\[
%D_g(X) = [(6x_1 - x^2_1) + (7x_2 - 1.5x^2_2) + (8x_3 - 0.5x^2_3)] / 49.17
%\]
In addition, the amount of water available for allocations is not precisely defined either. It is more or less about $6$ units of water, which is a {\em fuzzy constraint} that is defined by the fuzzy membership function:
\[
\begin{array}{lcl}
D_c(X) = 1 \; & if & \; x_1 + x_2 + x_3 \leq 5 \\
D_c(X) = \frac{7- (x_1 + x_2 + x_3)}{ 2}  \; & if & \; 5 \leq  x_1 + x_2 + x_3 \leq 7  \\
D_c(X) = 0 \; & if & \; x_1 + x_2 + x_3 \geq 7
\end{array}
\]
In this fuzzy environment optimization problem, the target turns to maximize the degree of the total benefits, $T(X)$, having that the total amount of available water is more or less $6$ units of water, since the higher the value of $T(X)$, whose $x_1 + x_2 + x_3$ from $X$ is within the vicinity of $6$, the higher the degree of $T(X)$. Thus, this fuzzy optimization problem becomes:
\[
maximize \; minimum (D_g(X), D_c(X))
\]
subject to
\[
\begin{array}{lcl}
D_g(X) = \frac{(6x_1 - x^2_1) + (7x_2 - 1.5x^2_2) + (8x_3 - 0.5x^2_3)} {49.17} \\
D_c(X) = \frac{7- (x_1 + x_2 + x_3)}{ 2}
\end{array}
\]
The optimal fuzzy solution of this fuzzy water allocation optimization problem is $x_1 = 0.91$, $x_2 = 0.94$, $x_3 = 3.81$, $D_g(X) = 0.67$, and $D_c(X) = 0.67$ and with total benefits $T(X) = 33.1$, where $X = (x_1,x_2,x_3)$.
\label{ex:water}
\end{example}
To represent this fuzzy optimization problem in fuzzy answer set optimization and to provide correct solution to the problem, the fuzzy answer set optimization representation of the problem has to be able to represent the fuzzy membership function of the objective function (objective membership function) and the fuzzy membership function of the problem constraints (the fuzzy constraints) along with the preference relation that maximizes the minimum of both fuzzy membership functions, and to be able to compare for the maximum of the minimum of both membership functions across the generated fuzzy answer sets.

However, the current syntax and semantics of fuzzy answer set optimization do not define fuzzy preference relations or rank fuzzy answer sets based on minimization or maximization of some desired criterion specified by the user. Therefore, in this paper we extend fuzzy answer set optimization with fuzzy aggregate preferences to allow the ability to represent and reason and intuitively solve fuzzy optimization problems. Fuzzy aggregates fuzzy answer set optimization framework modifies and generalizes the classical aggregates classical answer set optimization presented in \cite{Saad_ASOG} as well as the classical answer set optimization introduced in \cite{ASO}. We show the application of fuzzy aggregates fuzzy answer set optimization to the fuzzy water allocation problem described in Example (\ref{ex:water}), where a fuzzy answer set program \cite{Saad_DFLP} (disjunctive fuzzy logic program with fuzzy answer set semantics) is used as fuzzy answer sets generator rules.

The framework of fuzzy aggregates fuzzy answer set optimization is built upon both the fuzzy answer set optimization programming \cite{Saad_FuzzPref} and the fuzzy answer set programming with fuzzy aggregates \cite{Saad_FuzzAgg}.

\section{Fuzzy Aggregates Fuzzy Answer Set Optimization}

Fuzzy aggregates fuzzy answer set optimization programs are fuzzy logic programs under the fuzzy answer set semantics whose fuzzy answer sets are ranked according to fuzzy preference relations specified in the programs. A fuzzy aggregates fuzzy answer set optimization program is a union of two sets of fuzzy logic rules, $\Pi =  R_{gen} \cup R_{pref}$. The first set of fuzzy logic rules, $R_{gen}$, is called the generator rules that generate the fuzzy answer sets that satisfy every fuzzy logic rule in $R_{gen}$. $R_{gen}$ is any set of fuzzy logic rules with well-defined fuzzy answer set semantics including normal, extended, and disjunctive fuzzy logic rules \cite{Saad_DFLP,Saad_eflp,Subrahmanian_B}, as well as fuzzy logic rules with fuzzy aggregates \cite{Saad_FuzzAgg} (all are forms of {\em fuzzy answer set programming}). The second set of fuzzy logic rules, $R_{pref}$, is called the {\em fuzzy preference rules}, which are fuzzy logic rules that represent the required {\em fuzzy quantitative} and {\em qualitative} preferences over the fuzzy answer sets generated by $R_{gen}$. The fuzzy preferences rules in $R_{pref}$ are used to rank the generated fuzzy answer sets from $R_{gen}$ from the top preferred fuzzy answer set to the least preferred fuzzy answer set. An advantage of fuzzy answer set optimization programs is that $R_{gen}$ and $R_{pref}$ are independent. This makes fuzzy preference elicitation easier and the whole approach is more intuitive and easy to use in practice. The syntax and semantics of fuzzy aggregates fuzzy answer set optimization programs are built on top of the syntax and semantics of both the fuzzy answer set optimization programs \cite{Saad_FuzzPref} and the fuzzy answer set with fuzzy aggregates programs \cite{Saad_FuzzAgg}.

\subsection{Basic Language}

Let ${\cal L}$ be a first-order language with finitely many predicate symbols, function symbols, constants, and
infinitely many variables. A term is a constant, a variable or a function. A literal is either an atom, $a$, in ${\cal L}$ or the negation of $a$, denoted by $\neg a$, where ${\cal B_L}$ is the Herbrand base of ${\cal L}$ and $\neg$ is the classical negation. Non-monotonic negation or the negation as failure is denoted by $not$. The Herbrand universe of $\cal L$ is denoted by $U_{\cal L}$. Let $Lit$ be the set of all literals in ${\cal L}$, where $Lit = \{a | a \in {\cal B_L} \} \cup \{\neg a | a \in {\cal B_L}\}$. Grade membership values are assigned to literals in $Lit$ as values from $[0,1]$. The set $[0,1]$ and the relation $\leq$ form a complete lattice, where the join ($\oplus$) operation is defined as $\alpha_1 \oplus \alpha_2 = \max (\alpha_1,\alpha_2)$ and the meet ($\otimes$) is defined as $\alpha_1 \otimes \alpha_2 = \min (\alpha_1,\alpha_2)$.

A \emph{fuzzy annotation}, $\mu$, is either a constant in $[0, 1]$ (called \emph{fuzzy annotation constant}), a variable ranging over $[0, 1]$ (called \emph{fuzzy annotation variable}), or $f(\alpha_1,\ldots,\,\alpha_n)$ (called \emph{fuzzy annotation function}) where $f$ is a representation of a monotone, antimonotone, or nonmonotone total or partial function $f: ([0, 1])^n \rightarrow [0, 1]$ and $\alpha_1,\ldots,\alpha_n$ are fuzzy annotations. If $l$ is literal and $\mu$ is a fuzzy annotation then $l:\mu$ is called a fuzzy annotated literal.

A symbolic fuzzy set is an expression of the form $\{ X : U \; | \; C \}$, where $X$ is a variable or a function term and $U$ is fuzzy annotation variable or fuzzy annotation function, and $C$ is a conjunction of fuzzy annotated literals. A ground fuzzy set is a set of pairs of the form $\langle x : u \; | \; C^g \rangle$ such that $x$ is a constant term and $u$ is fuzzy annotation constant, and $C^g$ is a ground conjunction of fuzzy annotated literals. A symbolic fuzzy set or ground fuzzy set is called a fuzzy set term. Let $f$ be a fuzzy aggregate function symbol and $S$ be a fuzzy set term, then $f(S)$ is said a fuzzy aggregate, where $f \in \{$$sum_F$, $times_F$, $min_F$, $max_F$, $count_F$$\}$. If $f(S)$ is a fuzzy aggregate and $T$ is a constant, a variable or a function term, called {\em guard}, then we say $f(S) \prec T$ is a fuzzy aggregate atom, where $\prec \in \{=, \neq, <, >, \leq, \geq \}$.

A {\em fuzzy optimization aggregate} is an expression of the form $max_\mu (f(S))$, $min_\mu (f(S))$, $max_x (f(S))$, $min_x (f(S))$, $max_{x \mu} (f(S))$, and $min_{x \mu} (f(S))$, where $f$ is a fuzzy aggregate function symbol and $S$ is a fuzzy set term.

\subsection{Fuzzy Preference Rules Syntax}

Let $A$ be a set of fuzzy annotated literals, fuzzy annotated fuzzy aggregate atoms, and fuzzy optimization aggregates. A boolean combination over $A$ is a boolean formula over fuzzy annotated literals, fuzzy annotated fuzzy aggregate atoms, and fuzzy optimization aggregates in $A$ constructed by conjunction, disjunction, and non-monotonic negation ($not$), where non-monotonic negation is combined only with fuzzy annotated literals and fuzzy annotated fuzzy aggregate atoms.

\begin{definition} Let $A$ be a set of fuzzy annotated literals, fuzzy annotated fuzzy aggregate atoms, and fuzzy optimization aggregates. A fuzzy preference rule, $r$, over $A$ is an expression of the form
\begin{eqnarray}
C_1 \succ C_2 \succ \ldots \succ C_k \leftarrow l_{k+1}:\mu_{k+1},\ldots, l_m:\mu_m, \notag \\ not\; l_{m+1}:\mu_{m+1},\ldots, not\;l_{n}:\mu_{n} \label{rule:pref}
\end{eqnarray}
where $l_{k+1}:\mu_{k+1}, \ldots, l_{n}:\mu_{n}$ are fuzzy annotated literals or fuzzy annotated fuzzy aggregate atoms and $C_1, C_2, \ldots, C_k$ are boolean combinations over $A$.
\end{definition}
Let $r$ be a fuzzy preference rule of the form (\ref{rule:pref}), $head(r) = C_1 \succ C_2 \succ \ldots \succ C_k$, and $body(r) = l_{k+1}:\mu_{k+1},\ldots, l_m:\mu_m, not\; l_{m+1}:\mu_{m+1},\ldots, not\;l_{n}:\mu_{n}$. Intuitively, a fuzzy preference rule, $r$, of the form (\ref{rule:pref}) means that any fuzzy answer set that satisfies $body(r)$ and $C_1$ is preferred over the fuzzy answer sets that satisfy $body(r)$, some $C_i$ $(2 \leq i \leq k)$, but not $C_1$, and any fuzzy answer set that satisfies $body(r)$ and $C_2$ is preferred over fuzzy answer sets that satisfy $body(r)$, some $C_i$ $(3 \leq i \leq k)$, but neither $C_1$ nor $C_2$, etc.

Let $f(S)$ be a fuzzy aggregate. A variable, $X$, is a local variable to $f(S)$ if and only if $X$ appears in $S$ and $X$ does not appear in the fuzzy preference rule that contains $f(S)$. A global variable is a variable that is not a local variable. Therefore, the {\em ground instantiation} of a symbolic fuzzy set $$S = \{ X : U  \; | \; C \}$$ is the set of all ground pairs of the form \\ $\langle \theta(X) : \theta (U) \; | \;  \theta  (C) \rangle$, where $\theta$ is a substitution of every local variable appearing in $S$ to a constant from $U_{\cal L}$. A ground instantiation of a fuzzy preference rule, $r$, is the replacement of each global variable appearing in $r$ to a constant from $U_{\cal L}$, then followed by the ground instantiation of every symbolic fuzzy set, $S$, appearing in $r$. The ground instantiation of a fuzzy aggregates fuzzy answer set optimization program, $\Pi$, is the set of all possible ground instantiations of every fuzzy rule in $\Pi$.

\begin{definition} Formally, a fuzzy aggregates fuzzy answer set optimization program is a union of two sets of fuzzy logic rules, $\Pi =  R_{gen} \cup R_{pref}$, where $R_{gen}$ is a set of fuzzy logic rules with fuzzy answer set semantics, the {\em generator} rules, and $R_{pref}$ is a set of fuzzy preference rules.
\end{definition}

\begin{example} The fuzzy water allocation optimization problem presented in Example (\ref{ex:water}) can be represented as as a fuzzy aggregates  fuzzy answer set optimization program $\Pi = R_{gen} \cup R_{pref}$, where $R_{gen}$ is a set of disjunctive fuzzy logic rules with fuzzy answer set semantics \cite{Saad_DFLP} of the form:
\[
\begin{array}{r}
domX_1(0.91) \vee domX_1(1)  \vee  domX_1(2)  \vee  domX_1(3) \vee \\ domX_1(4) \vee  domX_1(5) \vee domX_1(6) \vee domX_1(7). \\ \\
domX_2(0.94) \vee domX_2(1)  \vee  domX_2(2)  \vee   domX_2(3) \vee \\ domX_2(4) \vee  domX_2(5) \vee domX_2(6) \vee domX_2(7). \\ \\
domX_3(1) \vee domX_3(2)  \vee  domX_3(3)  \vee  domX_3(3.81) \vee  \\ domX_3(4)  \vee  domX_3(5) \vee domX_3(6) \vee domX_3(7).
\end{array}
\]

\[
\begin{array}{c}
firm_1(X, 6*X - X*X)  \leftarrow  domX_1(X). \\
firm_2(X, 7*X - 1.5*X*X)  \leftarrow  domX_2(X). \\
firm_3(X, 8*X - 0.5*X*X)  \leftarrow  domX_3(X). \\
objective(X_1,X_2,X_3, y): \frac{B_1 + B_2 + B_3}{49.17}  \leftarrow  firm_1(X_1, B_1), \\ firm_2(X_2, B_2),    firm_3(X_3, B_3). \\
%constr(X_1,X_2,X_3): 1
%
constr(X_1,X_2,X_3, y): \frac{7- (X_1 + X_2 + X_3)}{ 2}   \leftarrow  domX_1(X_1),  \\ domX_2(X_2), domX_3(X_3),
5 \leq X_1 + X_2 + X_3 \leq 7. \\
\leftarrow  domX_1(X_1), domX_2(X_2), domX_3(X_3),   X_1 + X_2 + X_3 \leq 5. \\
\leftarrow domX_1(X_1), domX_2(X_2), domX_3(X_3),   X_1 + X_2 + X_3 \geq 7.
\end{array}
\]
where $domX_1(X_1)$, $domX_2(X_2)$, $domX_3(X_3)$ are predicates represent the domains of possible values for the variables $X_1$, $X_2$, $X_3$ that represent the units of water allocations to firms one, two and three respectively, $firm_i(X_i, B_i)$ is a predicate that represents the amounts of benefits, $B_i$, that firm, $i$, gets after allocated, $X_i$, units of water, for $1 \leq i \leq 3$, $objective(X_1,X_2,X_3, y):f(B_1,B_2,B_3)$ is a fuzzy annotated predicate that represents the objective membership value, $f(B_1,B_2,B_3)$, for the assignments of units of water to the variables $X_1$, $X_2$, $X_3$, where $y$ is a dummy constant to encode the vector of constant values $X = (x_1, x_2, x_3)$, and $constr(X_1,X_2,X_3, y): f(X_1,X_2,X_3)$ is a fuzzy annotated predicate that represents the fuzzy constraint membership value, $f(X_1,X_2,X_3)$, for the assignments of units of water to the variables $X_1$, $X_2$, $X_3$, where $y$ is a dummy constant to encode the vector of constant values $X = (x_1, x_2, x_3)$.
\\

The set of fuzzy preference rules, $R_{pref}$, of $\Pi$ consists of the fuzzy preference rule
\[
\begin{array}{r}
max_\mu \{Y: min(V_1,V_2) \; | \; objective(X_1,X_2,X_3, Y):V_1, \\ constr(X_1,X_2,X_3, Y):V_2\} \leftarrow
\end{array}
\]
\label{ex:water-code}
\end{example}

\section{Fuzzy Aggregates Fuzzy Answer Set Optimization Semantics}

Let $\mathbb{X}$ denotes a set of objects. Then, we use $2^\mathbb{X}$ to denote the set of all {\em multisets} over elements in $\mathbb{X}$. Let $\mathbb{R}$ denotes the set of all real numbers and $\mathbb{N}$ denotes the set of all natural numbers, and $U_{\cal L}$ denotes the Herbrand universe. Let $\bot$ be a symbol that does not occur in ${\cal L}$. Therefore, the semantics of the fuzzy aggregates are defined by the mappings: $sum_F : 2^{\mathbb{R} \times [0, 1] } \rightarrow \mathbb{R} \times [0, 1]$, $times_F: 2^{\mathbb{R} \times [0, 1] } \rightarrow \mathbb{R} \times [0, 1]$, $min_F : (2^{\mathbb{R} \times [0, 1] } - \emptyset) \rightarrow \mathbb{R} \times [0, 1]$, $max_F: (2^{\mathbb{R} \times [0, 1] } - \emptyset) \rightarrow \mathbb{R} \times [0, 1]$, $count_F : 2^{U_{\cal L} \times [0, 1]}  \rightarrow \mathbb{N} \times [0, 1]$. The application of $sum_F$ and $times_F$ on the empty multiset return $(0,1)$ and $(1,1)$ respectively. The application of $count_F$ on the empty multiset returns $(0,1)$. However, the application of $max_F$ and $min_F$  on the empty multiset is undefined.

The semantics of fuzzy aggregates and fuzzy optimization aggregates in fuzzy aggregates fuzzy answer set optimization is defined with respect to a fuzzy answer set, which is, in general, a total or partial mapping, $I$, from $Lit$ to $[0,1]$. In addition, the semantics of fuzzy optimization aggregates $max_\mu (f(S))$, $min_\mu (f(S))$, $max_x (f(S))$, $min_x (f(S))$, $max_{x \mu} (f(S))$, and $min_{x \mu} (f(S))$ are based on the semantics of the fuzzy aggregates $f(S)$. We say, a fuzzy annotated literal, $l:\mu$, is true (satisfied) with respect to a fuzzy answer set, $I$, if and only if $\mu \leq I(l)$. The negation of a fuzzy annotated literal, $not \; l:\mu$, is true (satisfied) with respect to $I$ if and only if $\mu \nleq I(l)$ or $l$ is undefined in $I$. The evaluation of fuzzy aggregates and the truth valuation of fuzzy aggregate atoms with respect to fuzzy answer sets are given as follows. Let $f(S)$ be a ground fuzzy aggregate and $I$ be a fuzzy answer set. In addition, let $S_I$ be the multiset constructed from elements in $S$, where $S_I = \{\!\!\{ x : u  \; | \; \langle x : u \; | \; C^g \rangle \in S \wedge$ $C^g$ is true w.r.t. $I \}\!\!\}$. Then, the evaluation of $f(S)$ with respect to $I$ is, $f(S_I)$, the result of the application of $f$ to $S_I$, where $f(S_I) = \bot$ if $S_I$ is not in the domain of $f$ and
\begin{itemize}

\item $sum_F(S_I) = (\sum_{\: x : u \in S_I} \; x \; , \;  \min_{\: x : x  \in S_I} \; u ) $

\item $times_F(S_I) = (\prod_{\: x : u \in S_I} \; x \; , \;  \min_{\: x : u  \in S_I} \; u ) $

\item $min_F (S_I)= (\min_{\: x : u \in S_I} \; x \; , \;  \min_{\: x : u  \in S_I} \; u ) $

\item $max_F (S_I)= (\max_{\: x : u \in S_I} \; x \; , \;  \min_{\: x : u  \in S_I} \; u ) $

\item $count_F (S_I)= (count_{\: x : u \in S_I} \; x \; , \;  \min_{\: x : u  \in S_I} \; u )$

\end{itemize}

\subsection{Fuzzy Preference Rules Semantics}

In this section, we define the notion of satisfaction of fuzzy preference rules with respect to fuzzy answer sets.

Let $\Pi = R_{gen} \cup R_{pref}$ be a ground fuzzy aggregates fuzzy answer set optimization program, $I,I'$ be fuzzy answer sets of $R_{gen}$ (possibly partial), and $r$ be a fuzzy preference rule in $R_{pref}$. Then the satisfaction of a boolean combination, $C$, appearing in $head(r)$, by $I$ is defined inductively as follows:

\begin{enumerate}

\item $I$ satisfies $l:\mu$ iff  $\mu \leq I(l)$.

\item $I$ satisfies $not\;l:\mu$ iff $\mu \nleq I(l)$ or $l$ is undefined in $I$.

\item $I$ satisfies $f(S) \prec T : \mu$ iff $f(S_I) = (x, \nu) \neq \bot$ and $x \prec T$ and $\mu \leq \nu$.

\item $I$ satisfies $not \; f(S) \prec T :\mu $ iff $f(S_I) =  \bot$ or $f(S_I) = (x, \nu) \neq \bot$ and $x \nprec T$ or $\mu \nleq \nu$.

\item $I$ satisfies $max_\mu (f(S))$ iff $f(S_I) = (x, \nu) \neq \bot$ and for any $I'$, $f(S_{I'}) = (x', \nu') \neq \bot$ and $\nu' \leq \nu$ or $f(S_I) \neq \bot$ and $f(S_{I'}) = \bot$.

\item $I$ satisfies $min_\mu (f(S))$ iff $f(S_I) = (x, \nu) \neq \bot$ and for any $I'$, $f(S_{I'}) = (x', \nu') \neq \bot$ and $\nu \leq \nu'$ or $f(S_I) \neq \bot$ and $f(S_{I'}) = \bot$.

\item $I$ satisfies $max_x (f(S))$ iff $f(S_I) = (x, \nu) \neq \bot$ and for any $I'$, $f(S_{I'}) = (x', \nu') \neq \bot$ and $x' \leq x$ or $f(S_I) \neq \bot$ and $f(S_{I'}) = \bot$.

\item $I$ satisfies $min_x (f(S))$ iff $f(S_I) = (x, \nu) \neq \bot$ and for any $I'$, $f(S_{I'}) = (x', \nu') \neq \bot$ and $x \leq x'$ or $f(S_I) \neq \bot$ and $f(S_{I'}) = \bot$.

\item $I$ satisfies $max_{x \mu} (f(S))$ iff $f(S_I) = (x, \nu) \neq \bot$ and for any $I'$, $f(S_{I'}) = (x', \nu') \neq \bot$ and $x' \leq x$ and $\nu' \leq \nu$ or $f(S_I) \neq \bot$ and $f(S_{I'}) = \bot$.

\item $I$ satisfies $min_{x \mu} (f(S))$ iff $f(S_I) = (x, \nu) \neq \bot$ and for any $I'$, $f(S_{I'}) = (x', \nu') \neq \bot$ and $x \leq x'$ and $\nu \leq \nu'$ or $f(S_I) \neq \bot$ and $f(S_{I'}) = \bot$.

\item $I$ satisfies $C_1 \wedge C_2$ iff $I \models C_1$ and $I \models C_2$.

\item $I$ satisfies $C_1 \vee C_2$ iff $I \models C_1$ or $I \models C_2$.
\end{enumerate}
\label{def:satisfaction}
The satisfaction of $body(r)$ by $h$ is defined inductively as:
\begin{itemize}

\item $I$ satisfies $l_i:\mu_i$ iff $\mu_i \leq I(l_i)$

\item $I$ satisfies $not\;l_j:\mu_j$ iff $\mu_j \nleq I(l_j)$ or $l_j$ is undefined in $I$.

\item $I$ satisfies $f(S) \prec T : \mu$ iff $f(S_I) = (x, \nu) \neq \bot$ and $x \prec T$ and $\mu \leq \nu$.

\item $I$ satisfies $not \; f(S) \prec T :\mu $ iff $f(S_I) =  \bot$ or $f(S_I) = (x, \nu) \neq \bot$ and $x \nprec T$ or $\mu \nleq \nu$.

\item $I$ satisfies $body(r)$ iff $\forall(k+1 \leq i \leq m)$, $I$ satisfies $l_i : \mu_i$ and $\forall(m+1 \leq j \leq n)$, $I$ satisfies $not\; l_j : \mu_j$.
\end{itemize}
The application of any fuzzy aggregate, $f$, except $count_F$, on a singleton $\{x:u \}$, returns $(x, u)$, i.e., $f(\{x:u\}) = (x,u)$. Therefore, we use $max_\mu (S)$, $min_\mu (S)$ $max_x (S)$, $min_x (S)$, $max_{x \mu} (S)$, and $min_{x \mu} (S)$ as abbreviations for the fuzzy optimization aggregates $max_\mu (f(S))$, $min_\mu(f(S))$, $max_x (f(S))$, $min_x(f(S))$, $max_{x \mu}(f(S))$, and $min_{x \mu}(f(S))$ respectively, whenever $S$ is a singleton and $f$ is arbitrary fuzzy aggregate except $count_F$.
\begin{definition} Let $\Pi = R_{gen} \cup R_{pref}$ be a ground fuzzy aggregates fuzzy answer set optimization program, $I$ be a fuzzy answer set of $R_{gen}$, $r$ be a fuzzy preference rule in $R_{pref}$, and $C_i$ be a boolean combination in $head(r)$. Then, we define the following notions of satisfaction of $r$ by $I$:

\begin{itemize}
\item $I \models_{i} r$ iff $I \models body(r)$ and $I \models C_i$.

\item $I \models_{irr} r$ iff $I \models body(r)$ and $I$ does not satisfy any $C_i$ in $head(r)$.

\item $I \models_{irr} r$ iff $I$ does not satisfy $body(r)$.
\end{itemize}
\end{definition}
$I \models_{i} r$ means that $I$ satisfies the body of $r$ and the boolean combination $C_i$ that appears in the head of $r$. However, $I \models_{irr} r$ means that $I$ is irrelevant (denoted by $irr$) to $r$ or, in other words, $I$ does not satisfy the fuzzy preference rule $r$, because either one of two reasons. Either because of $I$ does not satisfy the body of $r$ and does not satisfy any of the boolean combinations that appear in the head of $r$. Or because $I$ does not satisfy the body of $r$.

\subsection{Fuzzy Answer Sets Ranking}

In this section we define the ranking of the fuzzy answer sets with respect to a boolean combination, a fuzzy preference rule, and with respect to a set of fuzzy preference rules.

Let $\Pi = R_{gen} \cup R_{pref}$ be a ground fuzzy aggregates fuzzy answer set optimization program, $I_1, I_2$ be two fuzzy answer sets of $R_{gen}$, $r$ be a fuzzy preference rule in $R_{pref}$, and $C_i$ be boolean combination appearing in $head(r)$. Then, $I_1$ is strictly preferred over $I_2$ w.r.t. $C_i$, denoted by $I_1 \succ_i I_2$, iff $I_1 \models C_i$ and $I_2 \nvDash C_i$ or $I_1 \models C_i$ and $I_2 \models C_i$ (except $C_i$ is a fuzzy optimization aggregate) and one of the following holds:

\begin{itemize}

\item $C_i = l:\mu$ implies $I_1 \succ_i I_2$ iff $I_1(l) > I_2(l)$.

\item $C_i = not \; l:\mu$ implies $I_1 \succ_i I_2$ iff $I_1(l) < I_2(l)$ or $l$ is undefined in $I_1$ but defined in $I_2$.

\item $C_i = f(S) \prec T : \mu$ implies $I_1 \succ_i I_2$ iff $f(S_{I_1}) = (x, \nu) \neq \bot$, $f(S_{I_2}) = (x', \nu') \neq \bot$, and $\nu' < \nu$.

    \item $C_i = not \; f(S) \prec T :\mu $ implies $I_1 \succ_i I_2$ iff

        \begin{itemize}
            \item $f(S_{I_1}) =  \bot$ and $f(S_{I_2}) \neq  \bot$ or
            \item $f(S_{I_1}) = (x, \nu) \neq \bot$, $f(S_{I_2}) = (x', \nu') \neq \bot$, and $\nu < \nu'$

        \end{itemize}

\item $C_i \in \{ max_\mu (f(S)), \; min_\mu (f(S)), \; max_x (f(S)), \\ min_x (f(S)),
          max_{x \mu} (f(S)), \; min_{x \mu} (f(S)) \}$ implies $I_1 \models C_i$ and $I_2 \nvDash C_i$.

\item $C_i = C_{i_1} \wedge C_{i_2}$ implies $I_1 \succ_i I_2$ iff there exists $t \in \{{i_1}, {i_2}\}$ such that $I_1 \succ_t I_2$ and for all other $t' \in \{{i_1}, {i_2}\}$, we have $I_1 \succeq_{t'} I_2$.

\item $C_i = C_{i_1} \vee C_{i_2}$ implies $I_1 \succ_i I_2$ iff there exists $t \in \{{i_1}, {i_2}\}$ such that $I_1 \succ_t I_2$ and for all other $t' \in \{{i_1}, {i_2}\}$, we have $I_1 \succeq_{t'} I_2$.

\end{itemize}
We say, $I_1$ and $I_2$ are equally preferred w.r.t. $C_i$, denoted by $I_1 =_{i} I_2$, iff $I_1 \nvDash C_i$ and $I_2 \nvDash C_i$ or $I_1 \models C_i$ and $I_2 \models C_i$ and one of the following holds:

\begin{itemize}

\item $C_i = l:\mu$ implies $I_1 =_{i} I_2$ iff $I_1(l) = I_2(l)$.

\item $C_i = not \; l:\mu$ implies $I_1 =_{i} I_2$  iff $I_1(l) = I_2(l)$ or $l$ is undefined in both $I_1$ and $I_2$.

\item $C_i = f(S) \prec T : \mu$ implies $I_1 =_{i} I_2$ iff $f(S_{I_1}) = (x, \nu) \neq \bot$, $f(S_{I_2}) = (x', \nu') \neq \bot$, and $\nu' = \nu$.

    \item $C_i = not \; f(S) \prec T :\mu $ implies $I_1 =_{i} I_2$ iff

        \begin{itemize}
            \item $f(S_{I_1}) =  \bot$ and $f(S_{I_2}) =  \bot$ or
            \item $f(S_{I_1}) = (x, \nu) \neq \bot$, $f(S_{I_2}) = (x', \nu') \neq \bot$, and $\nu = \nu'$

        \end{itemize}

\item $C_i \in \{ max_\mu (f(S)), \; min_\mu (f(S)), \; max_x (f(S)), \\ min_x (f(S)),
          max_{x \mu} (f(S)), \; min_{x \mu} (f(S)) \}$ implies $I_1 =_{i} I_2$ iff $I_1 \models C_i$ and $I_2 \models C_i$.

\item $C_i = C_{i_1} \wedge C_{i_2}$ implies $I_1 =_{i} I_2$ iff
\[\forall \: t \in \{{i_1}, {i_2}\}, \; I_1 =_{t} I_2. \]

\item $C_i = C_{i_1} \vee C_{i_2}$ implies $I_1 =_{i} I_2$ iff
\[
|\{I_1 \succeq_{t} I_2 \: | \: \forall \: t \in \{{i_1}, {i_2}\} \}| = | \{ I_2 \succeq_{t} I_1 \: | \: \forall \: t \in \{{i_1}, {i_2}\} \}|.
\]

%\item $C_i = C_{i_1} \vee C_{i_2}$ implies $I_1 =_{i} I_2$ iff $I_1 =_{i_1} I_2$ and \\ $I_1 =_{i_2} I_2$.
\end{itemize}
We say, $I_1$ is at least as preferred as $I_2$ w.r.t. $C_i$, denoted by $I_1 \succeq_i I_2$, iff $I_1 \succ_i I_2$ or $I_1 =_i I_2$.
\label{def:compination}

\begin{definition} Let $\Pi = R_{gen} \cup R_{pref}$ be a ground fuzzy aggregates fuzzy answer set optimization program, $I_1, I_2$ be two fuzzy answer sets of $R_{gen}$, $r$ be a fuzzy preference rule in $R_{pref}$, and $C_l$ be boolean combination appearing in $head(r)$. Then, $I_1$ is strictly preferred over $I_2$ w.r.t. $r$, denoted by $I_1 \succ_r I_2$, iff one of the following holds:
\begin{itemize}
\item $I_1 \models_{i} r$ and $I_2 \models_{j} r$ and $i < j$, \\
where $i = \min \{l \; | \; I_1 \models_l r \}$ and $j = \min \{l \; | \; I_2 \models_l r \}$.

\item $I_1 \models_{i} r$ and $I_2 \models_{i} r$ and $I_1 \succ_i I_2$, \\
where $i = \min \{l \; | \; I_1 \models_l r \} = \min \{l \; | \; I_2 \models_l r \}$.

\item $I_1 \models_{i} r$ and $I_2 \models_{irr} r$.
\end{itemize}
We say, $I_1$ and $I_2$ are equally preferred w.r.t. $r$, denoted by $I_1 =_{r} I_2$, iff one of the following holds:
\begin{itemize}
\item $I_1 \models_{i}  r$ and $I_2 \models_{i} r$ and $I_1 =_i I_2$, \\
where $i = \min \{l \; | \; I_1 \models_l r \} = \min \{l \; | \; I_2 \models_l r \}$.
\item $I_1 \models_{irr}  r$ and $I_2 \models_{irr} r$.
\end{itemize}
We say, $I_1$ is at least as preferred as $I_2$ w.r.t. $r$, denoted by $I_1 \succeq_{r} I_2$, iff $I_1 \succ_{r} I_2$ or $I_1 =_{r} I_2$.
\label{def:pref-rule}
\end{definition}
The above definitions specify how fuzzy answer sets are ranked according to a given boolean combination and according to a fuzzy preference rule. Definition \ref{def:compination} shows the ranking of fuzzy answer sets with respect to a boolean combination. However, Definition \ref{def:pref-rule} specifies the ranking of fuzzy answer sets according to a fuzzy preference rule. The following definitions determine the ranking of fuzzy answer sets with respect to a set of fuzzy preference rules.

\begin{definition} [Pareto Preference] Let $\Pi = R_{gen} \cup R_{pref}$ be a fuzzy aggregates fuzzy answer set optimization program and $I_1, I_2$ be fuzzy answer sets of $R_{gen}$. Then, $I_1$ is (Pareto) preferred over $I_2$ w.r.t. $R_{pref}$, denoted by $I_1 \succ_{R_{pref}} I_2$, iff there exists at least one fuzzy preference rule $r \in R_{pref}$ such that $I_1 \succ_{r} I_2$ and for every other rule $r' \in R_{pref}$, $I_1 \succeq_{r'} I_2$. We say, $I_1$ and $I_2$ are equally (Pareto) preferred w.r.t. $R_{pref}$, denoted by $I_1 =_{R_{pref}} I_2$, iff for all $r \in R_{pref}$, $I_1 =_{r} I_2$.
\end{definition}

\begin{definition} [Maximal Preference] Let $\Pi = R_{gen} \cup R_{pref}$ be a fuzzy aggregates fuzzy answer set optimization program and $I_1, I_2$ be fuzzy answer sets of $R_{gen}$. Then, $I_1$ is (Maximal) preferred over $I_2$ w.r.t. $R_{pref}$, denoted by $I_1 \succ_{R_{pref}} I_2$, iff
\[
|\{r \in R_{pref} | I_1 \succeq_{r} I_2\}| > |\{r \in R_{pref} | I_2 \succeq_{r} I_1\}|.
\]
We say, $I_1$ and $I_2$ are equally (Maximal) preferred w.r.t. $R_{pref}$, denoted by $I_1 =_{R_{pref}} I_2$, iff
\[
|\{r \in R_{pref} | I_1 \succeq_{r} I_2\}| = | \{r \in R_{pref} | I_2 \succeq_{r} I_1\}|.
\]
\end{definition}
Observe that the Maximal preference relation is more {\em general} than the Pareto preference relation, since the Maximal preference definition {\em subsumes} the Pareto preference relation.

\begin{example} The generator rules, $R_{gen}$, of the fuzzy aggregates fuzzy answer set program representation, $\Pi = R_{gen} \cup R_{pref}$, of the fuzzy water allocation optimization problem described in Example (\ref{ex:water-code}) has $38$ fuzzy answer sets, where the most relevant fuzzy answer sets with reasonably high grade membership values are:
{\small
\[
\begin{array}{l}
\hspace{-0.7cm} I_1 = \{ obj(4,0.94,1,y) :  0.42,  constr(4,0.94,1,y): 0.53, \ldots \} \\
\hspace{-0.7cm} I_2 = \{ obj(3,0.94,2,y):  0.57,  constr(3,0.94,2,y):  0.53, \ldots \} \\
\hspace{-0.7cm} I_3 = \{ obj(2,0.94,3,y) :  0.67,  constr(2,0.94,3,y) :  0.53, \ldots \} \\
\hspace{-0.7cm} I_4 = \{ obj(1,0.94,4,y) : 0.70,  constr(1,0.94,4,y) :  0.53, \ldots \} \\
\hspace{-0.7cm} I_5 = \{ obj(0.91,0.94,4,y) :  0.69,  constr(0.91,0.94,4,y) :  0.58, \ldots   \} \\
\hspace{-0.7cm} I_6 = \{ obj(1,0.94,3.81,y) :  0.68,  constr(1,0.94,3.81,y) :  0.63, \ldots \} \\
\hspace{-0.7cm} I_7 = \{ obj(0.91,0.94,3.81,y) :  0.67,  constr(0.91,0.94,3.81,y) :  0.67, \ldots \} \\
\hspace{-0.7cm} I_8 = \{ obj(0.91,1,3.81,y) :  0.68,  constr(0.91,1,3.81,y) :  0.64, \ldots \} \\
\hspace{-0.7cm} I_9 = \{ obj(1,1,3.81,y) :  0.69,  constr(1,1,3.81,y) :  0.60, \ldots \} \\
\hspace{-0.7cm} I_{10} = \{ obj(0.91,1,4,y) : 0.69,  constr(0.91,1,4,y) :  0.55, \ldots \} \\
\hspace{-0.7cm} I_{11} = \{ obj(0.91,2,3,y) : 0.65,  constr(0.91,2,3,y) :  0.55, \ldots \} \\
\hspace{-0.7cm} I_{12} = \{ obj(0.91,3,2,y) : 0.53,  constr(0.91,3,2,y) :  0.55, \ldots \} \\
\hspace{-0.7cm} I_{13} = \{ obj(0.91,4,1,y) :  0.33, constr(0.91,4,1,y) :  0.55, \ldots \}
\end{array}
\]
}
Notice that we use $obj(X_1,X_2,X_3, Y)$ instead of $objective(X_1,X_2,X_3, Y)$ for brevity. The ground instantiation of the fuzzy preference rule in $R_{pref}$ consists of one ground fuzzy preference rule, denoted by $r$, which is
{\small
\[
\begin{array}{l}
\hspace{-0.7cm} max_\mu \{
\\
\hspace{-0.7cm} \langle y: 0.42  |  obj(4,0.94,1,y) :  0.42,  constr(4,0.94,1,y) :  0.53  \rangle, \\
\hspace{-0.7cm} \langle y: 0.53  |  obj(3,0.94,2,y) :  0.57,  constr(3,0.94,2,y) :  0.53  \rangle, \\
\hspace{-0.7cm} \langle y: 0.53  |  obj(2,0.94,3,y) :  0.67,  constr(2,0.94,3,y) : 0.53  \rangle, \\
\hspace{-0.7cm} \langle y: 0.53  |  obj(1,0.94,4,y) :  0.70,  constr(1,0.94,4,y) :  0.53  \rangle, \\
\hspace{-0.7cm} \langle y: 0.58  |  obj(0.91,0.94,4,y) :  0.69,  constr(0.91,0.94,4,y) :  0.58  \rangle, \\
\hspace{-0.7cm} \langle y: 0.63  |  obj(1,0.94,3.81,y) :  0.68,  constr(1,0.94,3.81,y) :  0.63  \rangle, \\
\hspace{-0.7cm} \langle y: 0.67  |  obj(0.91,0.94,3.81,y) :  0.67,  constr(0.91,0.94,3.81,y) :  0.67  \rangle, \\
\hspace{-0.7cm} \langle y: 0.64  |  obj(0.91,1,3.81,y) :  0.68,  constr(0.91,1,3.81,y) :  0.64  \rangle, \\
\hspace{-0.7cm} \langle y: 0.60  |  obj(1,1,3.81,y) :  0.69,  constr(1,1,3.81,y) :  0.60  \rangle, \\
\hspace{-0.7cm} \langle y: 0.55  |  obj(0.91,1,4,y) :  0.69,  constr(0.91,1,4,y) :  0.55  \rangle, \\
\hspace{-0.7cm} \langle y: 0.55  |  obj(0.91,2,3,y) :  0.65,  constr(0.91,2,3,y) :  0.55  \rangle, \\
\hspace{-0.7cm} \langle y: 0.53  |  obj(0.91,3,2,y) :  0.53,  constr(0.91,3,2,y) :  0.55  \rangle, \\
\hspace{-0.7cm} \langle y: 0.33  |  obj(0.91,4,1,y) :  0.33,  constr(0.91,4,1,y) :  0.55 \rangle, \\
\ldots \}
\end{array}
\]
}
Therefore, it can be easily verified that $I_7 \models_1 r$ and
\[
\begin{array}{l}
I_1 \models_{irr} r,
I_2 \models_{irr} r,
I_3 \models_{irr} r,
I_4 \models_{irr} r,
I_5 \models_{irr} r, \\
I_6 \models_{irr} r,
I_8 \models_{irr} r,
I_9 \models_{irr} r,
I_{10} \models_{irr} r, \\
I_{11} \models_{irr} r,
I_{12} \models_{irr} r,
I_{13} \models_{irr} r
\end{array}
\]
This implies that $I_7$ is the top (Pareto and Maximal) preferred fuzzy answer set and represents the optimal fuzzy decisions of the fuzzy water allocation optimization problem described in Example (\ref{ex:water}). The fuzzy answer set $I_7$ assigns $0.91$ to $x_1$, $0.94$ to $x_2$, and $3.81$ to $x_3$ with grade membership value $0.67$ and with total benefits $33.1$, which coincides with the optimal fuzzy solution of the problem as described in Example (\ref{ex:water}).
\end{example}

\section{Properties}

In this section, we show that the fuzzy aggregates fuzzy answer set optimization programs syntax and semantics naturally subsume and generalize the syntax and semantics of classical aggregates classical answer set optimization programs \cite{Saad_ASOG} as well as naturally subsume and generalize the syntax and semantics of classical answer set optimization programs \cite{ASO} under the Pareto preference relation, since there is no notion of Maximal preference relation has been defined for the classical answer set optimization programs.

A classical aggregates classical answer set optimization program, $\Pi^c$, consists of two separate classical logic programs; a classical answer set program, $R^c_{gen}$, and a classical preference program, $R^c_{pref}$ \cite{Saad_ASOG}. The first classical logic program, $R^c_{gen}$, is used to generate the classical answer sets. The second classical logic program, $R^c_{pref}$, defines classical context-dependant preferences that are used to form a preference ordering among the classical answer sets of $R^c_{gen}$.
\\
\\
Any classical aggregates classical answer set optimization program, $\Pi^c = R^c_{gen} \cup R^c_{pref}$, can be represented as a fuzzy aggregates fuzzy answer set optimization program, $\Pi = R_{gen} \cup R_{pref}$, where all fuzzy annotations appearing in every fuzzy logic rule in $R_{gen}$ and all fuzzy annotations appearing in every fuzzy preference rule in $R_{pref}$ are equal to $1$, which means the truth value {\em true}. For example, for a classical aggregates classical answer set optimization program, $\Pi^c = R^c_{gen} \cup R^c_{pref}$, that is represented by the fuzzy aggregates fuzzy answer set optimization program, $\Pi = R_{gen} \cup R_{pref}$, the classical logic rule
\begin{eqnarray*}
a_1 \; \vee \ldots \vee \; a_k \leftarrow  a_{k+1}, \ldots, a_m, not\; a_{m+1},
\ldots, not\;a_{n}
\end{eqnarray*}
is in $R^c_{gen}$, where $\forall (1 \leq i \leq n)$, $a_i$ is an atom, iff
\begin{eqnarray*}
a_1:1 \; \vee \ldots \vee \; a_k:1 \leftarrow  a_{k+1}:1, \ldots, a_m:1, \notag \\ not\; a_{m+1}:1,
\ldots, not\;a_{n}:1
\end{eqnarray*}
is in $R_{gen}$. It is worth noting that the syntax and semantics of this class of fuzzy answer set programs are the same as the syntax and semantics of the classical answer set programs \cite{Saad_DFLP,Saad_eflp}. In addition, the classical preference rule
\begin{eqnarray}
C_1 \succ C_2 \succ \ldots \succ C_k \leftarrow l_{k+1},\ldots, l_m, not\; l_{m+1},\ldots, not\;l_{n}
\label{rule:classical-pref}
\end{eqnarray}
is in $R^c_{pref}$, where $l_{k+1}, \ldots, l_{n}$ are literals and classical aggregate atoms and $C_1, C_2, \ldots, C_k$ are boolean combinations over a set of literals, classical aggregate atoms, and classical optimization aggregates iff
\begin{eqnarray}
C_1 \succ C_2 \succ \ldots \succ C_k \leftarrow l_{k+1}:1,\ldots, l_m:1, \notag \\ not\; l_{m+1}:1,\ldots, not\;l_{n}:1
\label{rule:classical-fuzzy-pref}
\end{eqnarray}
is in $R_{pref}$, where $C_1, C_2, \ldots, C_k$ and $l_{k+1},\ldots, l_n$ in (\ref{rule:classical-fuzzy-pref}) are exactly the same as $C_1, C_2, \ldots, C_k$ and $l_{k+1},\ldots, l_n$ in (\ref{rule:classical-pref}) except that each classical aggregate appearing within a classical aggregate atom or a classical optimization aggregate in (\ref{rule:classical-fuzzy-pref}) involves a conjunction of literals each of which is associated with the fuzzy annotation $1$, where $1$ represents the truth value \emph{true}. In addition, any classical answer set optimization program is represented as a fuzzy aggregates fuzzy answer set optimization program by the same way as for classical aggregates classical answer set optimization programs except that classical answer set optimization programs  disallows classical aggregate atoms and classical optimization aggregates.

The following theorem shows that the syntax and semantics of fuzzy aggregates fuzzy answer set optimization programs subsume the syntax and semantics of the classical aggregates classical answer set optimization programs \cite{Saad_ASOG}.

\begin{theorem} Let $\Pi = R_{gen} \cup R_{pref}$ be the fuzzy aggregates fuzzy answer set optimization program equivalent to a classical aggregates classical answer set optimization program, $\Pi^c = R^c_{gen} \cup R^c_{pref}$. Then, the preference ordering of the fuzzy answer sets of $R_{gen}$ w.r.t. $R_{pref}$ coincides with the preference ordering of the classical answer sets of $R^c_{gen}$ w.r.t. $R^c_{pref}$ under both Maximal and Pareto preference relations.
\label{thm:theorem}
\end{theorem}
Assuming that \cite{ASO} assigns the lowest rank to the classical answer sets that do not satisfy either the body of a classical preference rule or the body of a classical preference and any of the boolean combinations appearing in the head of the classical preference rule, the following theorems show that the syntax and semantics of the fuzzy aggregates fuzzy answer set optimization programs subsume the syntax and semantics of the classical answer set optimization programs \cite{ASO}.

\begin{theorem} Let $\Pi = R_{gen} \cup R_{pref}$ be the fuzzy aggregates fuzzy answer set optimization program equivalent to a classical answer set optimization program, $\Pi^c = R^c_{gen} \cup R^c_{pref}$. Then, the preference ordering of the fuzzy answer sets of $R_{gen}$ w.r.t. $R_{pref}$ coincides with the preference ordering of the classical answer sets of $R^c_{gen}$ w.r.t. $R^c_{pref}$.
\label{thm:1}
\end{theorem}

\begin{theorem} Let $\Pi = R_{gen} \cup R_{pref}$ be a fuzzy aggregates fuzzy answer set optimization program equivalent to a classical answer set optimization program, $\Pi^c = R^c_{gen} \cup R^c_{pref}$. A fuzzy answer set $I$ of $R_{gen}$ is Pareto preferred fuzzy answer set w.r.t. $R_{pref}$ iff a classical answer set $I^c$ of $R^c_{gen}$, equivalent to $I$, is Pareto preferred classical answer set w.r.t. $R^c_{pref}$.
\label{thm:2}
\end{theorem}
Theorem \ref{thm:theorem} shows in general fuzzy aggregates fuzzy answer set optimization programs in addition can be used solely for representing and reasoning about multi objectives classical optimization problems by the classical answer set programming framework under both the Maximal and Pareto preference relations, by simply replacing any fuzzy annotation appearing in a fuzzy aggregates fuzzy answer set optimization program by the constant fuzzy annotation $1$. Furthermore, Theorem \ref{thm:1} shows in general that fuzzy aggregates fuzzy answer set optimization programs in addition can be used solely for representing and reasoning about qualitative preferences under the classical answer set programming framework, under both Maximal and Pareto preference relations, by simply replacing any fuzzy annotation appearing in a fuzzy aggregates fuzzy answer set optimization program by the constant fuzzy annotation $1$. Theorem \ref{thm:2} shows the subsumption result of the classical answer set optimization programs.

\section{Conclusions and Related Work}

We developed syntax and semantics of a logical framework for representing and reasoning about both quantitative and qualitative preferences in a unified logic programming framework, namely fuzzy answer set optimization programs. The proposed framework is necessary to allow representing and reasoning in the presence of both quantitative and qualitative preferences across fuzzy answer sets. This is to allow the ranking of fuzzy answer sets from the most (top) preferred fuzzy answer set to the least preferred fuzzy answer set, where the top preferred fuzzy answer set is the one that is most desirable.Fuzzy answer set optimization programs modify and generalize the classical answer set optimization programs described in \cite{ASO}. We have shown the application of fuzzy answer set optimization programs to the course scheduling problem with fuzzy preferences described in \cite{Saad_DFLP}

To the best of our knowledge, this development is the first to consider a logical framework for reasoning about quantitative preferences, in general, and reasoning about both quantitative and qualitative preferences in particular. However, qualitative preferences were introduced in classical answer set programming in various forms. In \cite{Schaub-Comp}, preferences are defined among the rules of the logic program, whereas preferences among the literals described by the logic programs are introduced in \cite{Sakama}. Answer set optimization (ASO) \cite{ASO} and logic programs with ordered disjunctions (LPOD) \cite{LPOD} are two answer set programming based preference handling approaches, where context-dependant preferences are defined among the literals specified by the logic programs. Application-dependant preference handling approaches for planning were presented in \cite{Son-Pref,Schaub-Pref07}. Here, preferences among actions, states, and trajectories are defined, which are based on temporal logic. The major difference between \cite{Son-Pref,Schaub-Pref07} and \cite{ASO,LPOD} is that the former are specifically developed for planning, but the latter are application-independent.

Contrary to the existing approaches for reasoning about preferences in answer set programming, where preference relations are specified among rules and literals in one program, an ASO program consists of two separate programs; an answer set program, $P_{gen}$, and a preference program, $P_{pref}$ \cite{ASO}. The first program, $P_{gen}$, is used to generate the answer sets, the range of possible solutions. The second program, $P_{pref}$, defines context-dependant preferences that are used to form a preference order among the answer sets of $P_{gen}$, and hence the preference
order among the set of possible solutions.

Following \cite{ASO}, fuzzy answer set optimization programs distinguish between fuzzy answer set generation, by $P_{gen}$, and fuzzy preference based fuzzy answer set evaluation, by $P_{pref}$, which has several advantages. In particular, $P_{pref}$ can be specified independently from the type of $P_{gen}$, which makes preference elicitation easier and the whole approach more intuitive and easy to use in practice. In addition, more expressive forms of fuzzy preferences can be represented in fuzzy answer set optimization programs, since they allow several forms of boolean combinations in the heads of preference rules.

In \cite{Saad_ASOG}, classical answer set optimization programs have been extended to allow aggregate preferences. The introduction of aggregate preferences to answer set optimization programs have made the encoding of general optimization problems and Nash equilibrium strategic games more intuitive and easy. The syntax and semantics of the classical answer set optimization programs with aggregate preference were based on the syntax and semantics of classical answer set optimization \cite{ASO} and aggregates in classical answer set programming \cite{Recur-aggr}. it has been shown in \cite{Saad_ASOG} that the syntax and semantics of classical answer set optimization programs with aggregate preferences subsumes the syntax and semantics of classical answer set optimization programs described in \cite{ASO}.

\bibliographystyle{named}
\bibliography{Saad13LFO}

\end{document}